\documentclass[11pt,draftcls, peerreview, onecolumn]{IEEEtran}

\usepackage{algorithm2e}

\usepackage{bbm}

\usepackage{rotating}
\usepackage{pifont}

\usepackage{cite}

\usepackage[]{graphicx}

\usepackage{bbm}

\usepackage[cmex10]{amsmath}
\usepackage{amsfonts}
\usepackage{amssymb}

\usepackage[font=footnotesize]{subfig}
\DeclareMathOperator*{\argmin}{\arg\!\min}

\begin{document}
%
\title{A Novel Semi-Supervised Algorithm for Rare Prescription Side Effect Discovery}
%
%
%

\author{Jenna~Reps,~\IEEEmembership{Student Member,~IEEE,}
        Jonathan~M.~Garibaldi,~\IEEEmembership{}
        Uwe~Aickelin,~\IEEEmembership{}
        Daniele~Soria,~\IEEEmembership{}
        Jack~E.~Gibson,~\IEEEmembership{}
        Richard~B.~Hubbard,~\IEEEmembership{}
\thanks{}
\thanks{}
\thanks{
}}

\maketitle
\IEEEpeerreviewmaketitle

\begin{abstract}
Drugs are frequently prescribed to patients with the aim of improving each patient's medical state, but an unfortunate consequence of most prescription drugs is the occurrence of undesirable side effects.  Side effects that occur in more than one in a thousand patients are likely to be signalled efficiently by current drug surveillance methods, however, these same methods may take decades before generating signals for rarer side effects, risking medical morbidity or mortality in patients prescribed the drug while the rare side effect is undiscovered.  In this paper we propose a novel computational meta-analysis framework for signalling rare side effects that integrates existing methods, knowledge from the web, metric learning and semi-supervised clustering.  The novel framework was able to signal many known rare and serious side effects for the selection of drugs investigated, such as tendon rupture when prescribed Ciprofloxacin or Levofloxacin, renal failure with Naproxen and depression associated with Rimonabant.  Furthermore, for the majority of the drug investigated it generated signals for rare side effects at a more stringent signalling threshold than existing methods and shows the potential to become a fundamental part of post marketing surveillance to detect rare side effects.   
   
\end{abstract}

\begin{IEEEkeywords}
Adverse Drug Reaction, THIN, Semi-Supervised Clustering, MUTARA, Observed Expected Ratio
\end{IEEEkeywords}

%
\IEEEpeerreviewmaketitle

\section{Introduction}
\label{sec:introduction}
%
%
%
%
\IEEEPARstart {N}{egative} side effects caused by prescribed medication currently present a huge burden for the healthcare service in terms of causing both patient morbidity or mortality and costing large sums of money \cite{Gautier2003} \cite{Patel2007} \cite{Davies2009}.  Investigations have shown that the rate of unwanted side effects has been increasing annually \cite{Hartholt2010} \cite{Shepherd2012}. Possible reasons for this are an increase in the number of annual prescriptions due to an aging population or an increase in polypharmacy, when numerous drugs are prescribed at the same time \cite{Betteridge2012}.  Although it is common for a patient to develop side effects due to prescribed medication there is currently no efficient means of identifying all the side effects of a drug.  When the side effect is detrimental to the patient's quality of life, it is often referred to as an Adverse Drug Event (ADE) and when the drug causing the ADE is known, it is termed an Adverse Drug Reaction (ADR). A study conducted in the UK between November 2001 to April 2002 indicated that 6.5\% of admissions to hospital were due to ADRs, with the mortality rate for an ADR patient of 2.3\% \cite{Pirmohamed2004}.  Interestingly, it was found that over 70\% of these ADRs were potentially avoidable. A more recent study in Brazil suggests ADRs may be the cause of an even higher proportion of hospital admissions for the elderly as it showed that ADRs were the cause of hospitalisation for over 50\% of elderly patients \cite{Varallo2011}.  It also highlighted that a significant factor for developing an ADR was polypharmacy \cite{Varallo2011}, when patients are prescribed multiple drugs.

Some obvious ADRs can be discovered during the experimental stages of a drug's development, but the occurrence of an ADR can depend on a magnitude of factors and it is impossible to investigate all the possible situations that may occur when the drug is taken. For example, testing for ADRs that result from polypharmacy would require clinical trials with millions of people to be able to investigate all the different drug combinations and this is not possible.  Due to the limitations of clinical trials, rare ADRs, including fatal ones, are in most circumstances not discovered before a drug is marketed \cite{Berlin2008} \cite{Seruga2011}.  As a consequence, after a drug is approved and available to patients, possible ADRs are investigated during the whole lifetime of a drug by a process known as post-marketing drug surveillance.

Post-marketing surveillance (such as doctors being vigilant and noticing possible drug and illness associations) can identify common ADRs and in general the more common the ADR is, the fewer the number of patients that need to be prescribed the drug before it is discovered.  However, ADRs that occur for drugs that are rarely prescribed or rare ADRs may go unnoticed by medical practitioners and may cause morbidity or mortality in patients that could have been prevented with more efficient drug surveillance methods.  For example, it took 23 years before there was sufficient evidence that the drug Tamoxifen used to treat breast cancer caused endometrial cancer in about 1 in 6000 patients \cite{Berman2000} \cite{Ladewski2003}.  

Current methods to discover rare ADRs often involve using a Spontaneous Reporting System (SRS) database that contains a collection of voluntary suspected drug and ADR reports, such as the database containing information from the UK yellow card scheme.  The algorithms that signal ADRs by mining SRS databases calculate a measure of how disproportionally more often the medical event is reported with a specific drug of interest compared with any drug.  The frequently implemented measures of disproportionality involve using standard epidemiology measures \cite{Evans2001}, estimating the information component using a neural network approach \cite{Bate1998} or calculating a modified version of the relative risk by applying a Bayesian model \cite{Dumouchel1999}. SRS databases combine reports of possible ADRs from a large population enabling the identification of possible ADR signals more efficiently, but they are known to suffer from under-reporting \cite{Hazell2006} and this causes a lag in the time it takes to confidently signal a potential ADR.  The under-reporting may also prevent the detection of rare ADRs, as these ADRs may never be suspected and therefore never be reported to an SRS database.

A potential new way to detect the rare ADRs that cannot be identified by doctors or by current methods applied to SRS databases is to use The Health Improvement Network (THIN) database (www.thin-uk.co.uk),  an Electronic Healthcare Database (EHD) containing complete UK General Practice records for registered patients.  The THIN database contains all medical events (such as illnesses, laboratory results, signs and symptoms or administrative events) that a doctor is informed of for a patient as well as their complete prescription histories.  Therefore, any rare ADRs that are serious enough to be reported to a doctor are more likely to be detected at an earlier point in time by applying a suitable data mining method on the THIN database rather than mining the SRS databases. 

Existing methods developed for the EHDs are often disproportionality based methods (methods that contrast how often the event of interest occurs after the specified drug relative to how often the event of interest occurs after any drug) similar to the SRS methods \cite{Noren2010} \cite{Schuemie2011} \cite{Zorych2011} or association rule mining methods \cite{Jin2006} \cite{Jin2010} \cite{Ji2012}.  As EHDs do not contain links between drugs and suspected ADRs these are often inferred by investigating medical events that occur within some time period around a drug, but many of these medical events are linked to the cause of taking the drug and these `therapeutically related' medical events present a major issue with the majority of the existing methods. It has been demonstrated that these existing methods are currently not suitable for signalling rare ADRs \cite{Reps2013}.  However, it is the rare ADRs that are unlikely to be detected by mining SRS databases, due to under-reporting, so developing an algorithm that can signal rare ADRs using the THIN database would be beneficial.  

In this paper we develop a novel computational meta-analysis framework that integrates the existing methods (MUTARA, HUNT, OE ratio, see section \ref{sec:existing}) and uses information obtained from the internet to efficiently and accurately identify rare ADRs that occur immediately or shortly after a drug is prescribed.  The framework uses the dependency measures obtained from some of the existing electronic healthcare based methods and novel values of interest as attributes for each medical event that occurs within a 30 day period after the drug of interest is prescribed for any patient.  After the attributes are generated for each medical event we label some of the medical events by extracting information from the internet informing us of the indicator events  and known ADRs for the drug of interest.  The unlabelled medical events then have labels assigned by applying metric learning and semi-supervised clustering.  Finally using the labels we develop a novel filter that removes medical events labelled as indicator events and then return the remaining medical events ordered by how often they occurred unpredictably within 30 days after the drug being investigated multiplied by weights based on their assigned labels.

The continuation of this paper is as follows, section \ref{sec:background} contains the background information on the THIN database and section \ref{sec:problem} describes the problem formulation.  This is followed by the description of the novel methodology we developed that is able to identify rare ADRs in section \ref{sec:dress}.  Section \ref{sec:results} contains the results of comparing our novel algorithm with a section of existing methods for the detection of known rare ADRs and the discussion of these results is contained in section \ref{sec:disc}.  The paper finishes with the conclusion and future work suggestions in section \ref{sec:con}.    

\hfill

\section{THIN Database}
\label{sec:background}
The THIN database contains complete medical and prescription records for registered patients at participating general practices within the UK.  The medical information is recorded into the THIN database by Read Codes that correspond to illnesses, so each Read Code is paired with the illness description.
\begin{figure}[!t]
\centering
\includegraphics[width=0.8\textwidth]{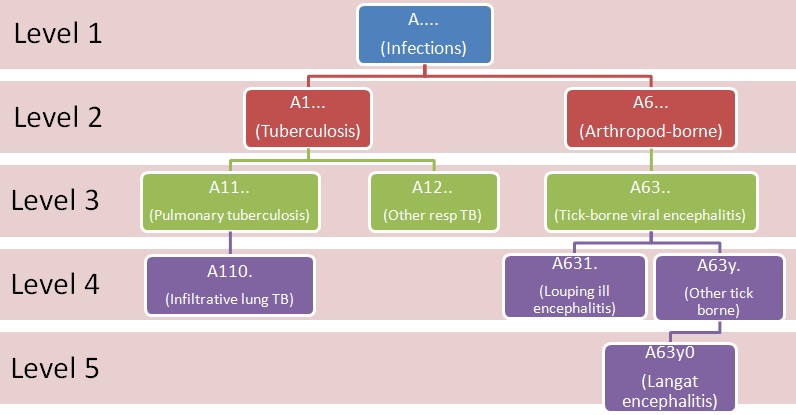}
\caption{Example of a branch in the Read Codes tree.}
\label{fig:rctree}
\end{figure}

Each Read Code is five elements long and the Read Codes have a tree structure. A level one Read Code has `.' as it's second element and corresponds to a very general description of an illness, for example `A....' is a level one Read Code that corresponds to `Infections'.  A level two Read Code has `.' for it's third element but not for it's second element, an example is `A1...' corresponding to `Tuberculosis'. A level two Read Code is the child of the level one Read Code with a matching first element, so `A1...' is the child of `A....' (or A....' is the parent of `A1...') and a child Read Code corresponds to a more specific version of it's parent's illness.  Fig (\ref{fig:rctree}) shows examples of different branches of the Read Code tree with the illness becoming more specific as the Read Code level increases. 

Each drug is recorded into the THIN database by a drugcode (sometimes called the multilexeid) that is paired to the generic name.  The drugcode consists of nine numbers and does not have a structure we use but the drugcode does specify the way the drug is ingested and the dosage.  Each entry also includes the date that a Read Code or drugcode is recorded but does not contain the time.  In this paper we used a subset of the THIN database containing records from 495 general practices. The subset contained approximately four million patients, over 358 million prescription entries and over 233 million medical event entries.

Patients can register at a new practice at any point over their lifetime and it has been shown that statistical studies using the THIN database will be biased if records from the first year after registration for a patient are included in the study \cite{Lewis2005}.  The reason for this is that newly registered patients will need to inform their new doctor of any chronic illnesses they have, but these illnesses will be recorded on the day they inform the doctor of them rather than the actual day they were discovered.  To prevent bias in this study we do not include the first year of a patient's medical records since registration.  We also ignore the last $30$ days of prescription records for each patient to reduce potential under reporting that may occur by including patients with less than 30 days of medical records after the first prescription of a drug.  

\begin{table}[t]
\centering
\caption{Examples of both a medical table and prescription table containing the patients records.  The column patID is a unique reference corresponding to a patient, the age is the patient's age in days when the entry was recorded and the Read Code/drugcode is the reference to the specific medical event or prescription respectively.}
\label{thin_ex}

\begin{tabular}{ccc|ccc}
\hline
\multicolumn{3}{c|}{Drug Table} & \multicolumn{3}{|c}{Medical Table} \\
\hline
patID & age & drugcode & patID & age & Read Code  \\
jj3 & 10000 & 979596759 & jj3 & 9999 & A123. \\
jj3 & 20000 & 969686881 &jj3 & 10000 & F1... \\
jj3 & 20001 & 969686881 &jj3 & 10000 & C1... \\
jj3 & 20001 & 912314611 &jj3 & 10001 & A123. \\
aa2 & 15001 & 912314611 &jj3 & 10013 & D25.. \\
aa2 & 15031 & 912314611 &jj3 & 18020 & C12.. \\
aa2 & 15061 & 912314611 &jj3 & 20001 & C121. \\
aa2 & 25304 & 912314611 &aa2 & 15001 & B21.. \\
bb8 & 10000 & 979596759 &bb8 & 9078 & A123. \\ 
&&&bb8 & 10000 & F12.. \\ 
&&&bb8 & 10000 & A123. \\ 
&&&bb8 & 10010 & D25.. \\ 
&&&bb8 & 27002 & C11.. \\ 
\end{tabular}
\end{table}    
\section{Problem Formulation}
\label{sec:problem}
The focus of this paper is to develop a method for detecting rare ADRs that occur shortly after prescription, so we can restrict our attention to the medical events that occur around the time of the drug prescription.  In this paper we consider the natural numbers $\mathbb{N}$ to include $0$ and use the interval $[a...b]$ to be the interval of natural numbers from $a$ to $b$ ($\{n \in \mathbb{N}: a \leq n \leq b \}$). 

If we let $M$ denote the set of Read Codes, $D$ denote the set of drugcodes and $\Omega$ denote the set of patients in the THIN database then the sequence of Read Codes or the sequence of drugcodes for a patient can be represented as two discrete functions.
The discrete function representing a patient's Read Code sequence is a mapping $f_{M}:\mathbb{N}\times \Omega \to \mathcal{P}(M); f_{M}(t,\omega) \to e_{\omega, t}$ where $t$ is the age in days of the patient $\omega$ and $e_{\omega,t}$ is the set of Read Codes that are recorded into the THIN database for patient $\omega$ when they are $t$ days old.  Similarly, the discrete function representing a patient's drugcode sequence is $f_{D}:\mathbb{N}\times \Omega \to \mathcal{P}(D); f_{D}(t, \omega) \to d_{\omega, t}$ where $d_{\omega, t}$ is the set of drugcodes recorded in THIN for the patient $\omega$ when they are $t$ days old.

Using the example THIN database entries shown in Table \ref{thin_ex}, we have,
\begin{equation}
\begin{split}
M&=\{ A123.,B21.,C1...,C11..,C12.., C121., D25..,F1...,F12.. \} \\ 
D&=\{ 979596759, 969686881, 912314611 \}\\
\Omega &= \{ jj3, aa2, bb8 \} \\
\end{split}
\end{equation}
The set of Read Codes that patient $jj3$ has recorded when he is $9999$, $10000$ and $10002$ days old are  $f_{M}(9999,jj3)=\{A123.\}$, $f_{M}(10000,jj3)=\{F1... ,C1...\}$ and  $f_{M}(10002,jj3)=\{\}=\emptyset$ respectively.  The reason $f_{M}(10002,jj3)=\{\}=\emptyset$ is that patient $jj3$ does not have a Read Code recorded when he is $10002$ days old.  Similarly, the set of drugcodes that patient $jj3$ has recorded when he is $10000$ and $10002$ days old are $f_{D}(10000,jj3)=\{ 979596759\}$ and $f_{D}(10002,jj3)=\{ \}=\emptyset$ respectively.

The set consisting of every age in days where patient $\omega$ has a drugcode recorded into the THIN database is,
\begin{equation}
A_{D}(\omega) = \{ t | f_{D}(t, \omega) \neq \emptyset\}
\end{equation}

The set of all drugs that are prescribed for patient $\omega$ is the finite union of the set of drugcodes recorded daily for the patient while they are active in the THIN database,  $\underset{t\in A_{D}(\omega)}{\cup f_{D}(t, \omega)}$, so to determine the age that patient $\omega$ is first prescribed the drug of interest $d_{*} \in D$ we first find the set of ages that the patient was prescribed the drug,
\begin{equation}
\alpha(\omega, d_{*})= \{t \in A_{D}(\omega)| d_{*} \in f_{D}(t, \omega) \}
\end{equation}  
and then define a new function $\alpha_{1}: \Omega \times D \to \mathbb{N} \cup \{-1\}:$
\begin{equation}
 \alpha_{1}(\omega, d_{*})= 
 \begin{cases}
 \min(\alpha(\omega, d_{*}))  & \mbox{if } d_{*} \in \underset{t\in A_{D}(\omega)}{\cup}f_{D}(t, \omega)\\
 -1 & \mbox{else}
 \end{cases}
\end{equation} 
that finds the minimum age that the patient is prescribed the drug or returns $-1$ if the patient has never been prescribed the drug.  The set of patient's ages where the drug is prescribed for the first time in 13 months is determined by the function $\hat{\alpha}: \Omega \times D \to \mathcal{P}(\mathbb{N} \cup \{-1\})$,    
\begin{equation}
 \hat{\alpha}(\omega, d_{*})= \{t \in \alpha(\omega, d_{*})| (t-s^{*})\geq 386, s^{*}= \argmin_{ \substack{s \in \alpha(\omega, d_{*}) \\ s <t}}(t-s)  \} \cup \alpha_{1}(\omega, d_{*}) 
\end{equation} 
If a patient is not prescribed the drug then the function returns $-1$ otherwise it returns the set of ages in days that the patient took the drug and had a minimum of 386 days between previously taking the drug. 

For the example entries in Table \ref{thin_ex}, the set of ages in days that patient $jj3$ is prescribed any drugs is $A_{D}(jj3)=\{10000, 20000, 20001 \}$ and the set of ages in days that the patient is prescribed the specific drugs $979596759$ and $969686881$ is $\alpha(jj3, 979596759)=\{10000 \}$ and $\alpha(jj3, 969686881)=\{20000, 20001 \}$ respectively.  The minimum age that patient $jj3$ is prescribed $969686881$ is $\alpha_{1}(jj3, 969686881)=20000$ but as the patient $aa2$ is never prescribed the drug $969686881$, $\alpha_{1}(aa2, 969686881)=-1$.  The patient $aa2$ is prescribed the drug $912314611$ four times ($\alpha(aa2, 912314611)=\{15001, 15031,15061, 25304 \}$) with the first prescription occurring at $15001$ days old ($\alpha_{1}(aa2, 912314611)=15001$) but the drug was then repeated monthly two times and then the patient had a $10243$ day break before being prescribed the drug for the final time, therefore the set of ages in days that patient $aa2$ is prescribed drug $912314611$ for the first time in 13 months is $\hat{\alpha} (aa2, 912314611)=\{15001, 25304 \}$ as the instances when the drug was repeated after $30$ days are not included.

In the continuation of this paper we will use $\hat{\alpha}(\omega, d_{*})_{k}$ to refer to the $k^{th}$ time patient $\omega$ is prescribed drug $d_{*}$ for the first time in 13 months, so $\hat{\alpha}(\omega, d_{*})_{1} < \hat{\alpha}(\omega, d_{*})_{2} <...<\hat{\alpha}(\omega, d_{*})_{n-1}<\hat{\alpha}(\omega, d_{*})_{n}$, where $n$ is the total number of times patient $\omega$ is prescribed drug $d_{*}$ for the first time in 13 months ($n=|\hat{\alpha}(\omega, d_{*})|$).  Following on with our example, $\hat{\alpha} (aa2, 912314611)_{1}=15001$ and $\hat{\alpha} (aa2, 912314611)_{2}=25304$. 

This then enables us to define a interval of interest around each time the patient is prescribed a drug for the first time in 13 months, for each $K \leq |\hat{\alpha}(\omega, d_{*})|$,
\begin{equation}
\label{eq:time_int}
T(\omega, d_{*}, t_{1}, t_{2})_{K}=
\begin{cases}
[(\hat{\alpha}(\omega, d_{*})_{K}+t_{1} )... (\hat{\alpha}(\omega, d_{*})_{K}+t_{2})] & \mbox{if } \hat{\alpha}(\omega, d_{*})_{K} \neq -1 \\
\emptyset & \mbox{else} \\
\end{cases} 
\end{equation}
Equation (\ref{eq:time_int}) is defining a time interval centred around the K$^{th}$ time drug $d_{*}$ is prescribed for the first time in 13 months for patient $\omega$ determine by the integers $t_{1}$ and $t_{2}$.  For example, if we wanted to investigate the 30 days after a prescription we would use $t_{1}=1$ and $t_{2}=30$ and the time period around the first prescriptions in 13 months of drug $912314611$ for patient $aa2$ would be $T(aa2, 912314611, 1, 30)_{1}=[(\hat{\alpha} (aa2, 912314611)_{1}+1)...(\hat{\alpha} (aa2, 912314611)_{1}+30)]=[(15001+1)...(15001+30)]=[(15002)...(15031)]$ and $T(aa2, 912314611, 1, 30)_{2}=[(\hat{\alpha} (aa2, 912314611)_{2}+1)...(\hat{\alpha} (aa2, 912314611)_{2}+30)]=[(25304+1)...(25304+30)]=[(25305)...(25334)]$.  If the patient is never prescribed the drug then the time period of interest is the empty set, for example $T(aa2, 979596759, 1, 30)_{1}=\emptyset$.  

As the function $f_{M}(t, \omega)$ returns a set of Read Codes that are recorded into the THIN database for patient $\omega$ when they are $t$ days old, we can find the set of ages in days that $\omega$ has any Read Code recorded,
\begin{equation}
A_{M}(\omega) = \{ t | f_{M}(t, \omega) \neq \emptyset\}
\end{equation} 
and the finite union over each $t \in A_{M}(\omega)$, $\underset{t \in A_{M}(\omega)}{\cup f_{M}(t, \omega)}$, is the set of all Read Codes that are recorded into the THIN database for the patient $\omega$.  It follows that $\underset{t \in T(\omega, d_{*}, t_{1}, t_{2})_{K}}{\cup f_{M}(t, \omega)} $ is the set of all Read Codes that are recorded into the THIN database for the patient $\omega$ during the period of interest determined by $t_{1}$ and $t_{2}$ around the $K^{th}$ prescription of drug $d_{*}$.  The function $h:M \times D \times \Omega \times \mathbb{Z}^{2} \to \{0,1\}$,
\begin{equation} 
h(e_{i}, d_{*}, \omega, t_{1}, t_{2}) =    \mathbbm{1}\{ e_{i} \in \underset{t \in T(\omega, d_{*}, t_{1}, t_{2})_{1}}{\cup} f_{M}(t, \omega) \} 
\end{equation}

is one if the patient $\omega$ has the Read Code $e_{i}$ recorded within the time period of interest around the first prescription of drug $d_{*}$ and zero otherwise.  To determine the number of times the drug $d_{*}$ is prescribed to patient $\omega$ for the first time in 13 months and the Read Code $e_{i}$ is recorded within the time period of interest we define another function $\hat{h}:M \times D \times \Omega \times \mathbb{Z}^{2} \to \mathbb{N}$,

\begin{equation} 
\hat{h}(e_{i}, d_{*}, \omega, t_{1}, t_{2}) = \sum_{K\leq |\hat{\alpha}(\omega, d_{*})|} {\mathbbm{1}\{ e_{i} \in \underset{t \in T(\omega, d_{*}, t_{1}, t_{2})_{K}}{\cup} f_{M}(t, \omega) \} }
\end{equation}

The total number of patients in the THIN database that have Read Code $e_{i}$ recorded within the time period of interest centred around the first prescription of drug $d_{*}$ is then, 
\begin{equation}
\sum_{\omega \in \Omega}{h(e_{i}, d_{*}, \omega, t_{1}, t_{2})}
\end{equation}  

and the total number of first times prescriptions of drug $d_{*}$ in 13 months where the Read Code $e_{i}$ is recorded within the time period centred around the prescription determined by $t_{1}$ and $t_{2}$ is,
\begin{equation}
\sum_{\omega \in \Omega}{\hat{h}(e_{i}, d_{*}, \omega, t_{1}, t_{2})}
\end{equation}  

\subsection{Existing Algorithms}
\label{sec:existing}
The observed to expected (OE) ratio \cite{Noren2010} calculates the information component ($IC$) that looks at the disproportionality between how often a Read Code is recorded within some time period after the first prescription in 13 months of the drug of interest compared to how often it is recorded within the same time period after the first prescription in 13 months of any drug but also adds a bias lowering the IC value if the Read Code or drug is rare.  

The function used to test if a set of medical events is empty is, $\hat{H}:\mathcal{P}(M) \to \{0,1\}$,
\begin{equation}
\hat{H}[B] =
\begin{cases}
0 & \mbox{ if } B=\emptyset\\
1 & \mbox{ else} \\
\end{cases}
\end{equation}
using the above function, we define the following values,
\begin{equation}
\begin{split}
&n_{d_{*} e_{i}}(t_{1}, t_{2}) =\sum_{\omega \in \Omega}{\hat{h}(e_{i}, d_{*}, \omega, t_{1}, t_{2})}\\
&n_{\bullet e_{i}}(t_{1}, t_{2}) =\sum_{d \in D} {\sum_{\omega \in \Omega}{\hat{h}(e_{i}, d, \omega, t_{1}, t_{2})}}\\
&n_{d_{*} \bullet}(t_{1}, t_{2}) = \sum_{\omega \in \Omega}{\overset{|\hat{\alpha}(\omega, d_{*})|}{\sum_{K=1}}{ \hat{H}[ M \cap (\underset{t \in T(\omega, d_{*}, t_{1}, t_{2})_{K}}{\cup f_{M}(t, \omega)})  ] }}\\
&n_{\bullet \bullet} (t_{1}, t_{2}) = \sum_{d \in D} \sum_{\omega \in \Omega}{\overset{|\hat{\alpha}(\omega, d)|}{\sum_{K=1}}{\hat{H}[ M \cap \underset{t \in T(\omega, d, t_{1}, t_{2})_{K}}{\cup f_{M}(t, \omega)}  ]}}
\end{split}
\end{equation}
Where $n_{d_{*} e_{i}}(0, 30)$ is the number of times that event $e_{i}$ occurs in the month after a first prescriptions in 13 months of drug $d_{*}$, $n_{.e_{i}}(0, 30)$ is the number of times that event $e_{i}$ occurs in the month after a first prescriptions in 13 months of any drug, $n_{d_{*}.}(0, 30)$ is the number of times a drug $d_{*}$ is prescribed for the first time in 13 months and has any follow up in the month after the prescription and $n_{..} (0, 30)$ is the number of times any drug is prescribed for the first time in 13 months and has any follow up in the month after the prescription. 

The expected number of first time prescriptions of drug $d_{*}$ in 13 months that have the Read Code $e_{i}$ occur within the time period defined by $t_{1}$ and $t_{2}$ if the Read Code occurs independently of the drug is,
\begin{equation}
E_{d_{*}e_{i}}(t_{1}, t_{2})=\frac{  n_{d_{*} \bullet}(t_{1}, t_{2})  n_{\bullet e_{i}}(t_{1}, t_{2})  }{ n_{\bullet \bullet}(t_{1}, t_{2})  }
\end{equation}
and the $IC(t_{1}, t_{2}, e_{i}, d_{*})$ value calculates how many first prescription in 13 months of drug $d_{*}$ have the Read Code $e_{i}$ occurring between $t_{1}$ and $t_{2}$ days after divided by the expected value if the Read Code and drug occur independently,
\begin{equation}
IC(t_{1}, t_{2}, e_{i}, d_{*})  = \log_{2}(\frac{n_{d_{*} e_{i}}(t_{1}, t_{2})  +1/2 }{   E_{d_{*} e_{i}}(t_{1}, t_{2})+1/2  }) 
\end{equation}
The value of a half added to both the numerator and denominator in the $IC(t_{1}, t_{2}, e_{i}, d_{*})$ value calculation creates a bias that causes the $IC(t_{1}, t_{2}, e_{i}, d_{*})$ value to tend to zero when the Read Code or drug occurs rarely.

The $IC_{\Delta}$ is a measure that compares the $IC(0, 30, e_{i}, d_{*})$ corresponding to the time period of a month after the first prescription in 13 months compared to the $IC(-810,-630, e_{i}, d_{*})$ corresponding to a time period between 27 and 21 months prior to the prescription.
\begin{equation}
\label{eq:icdelta}
IC_{\Delta}(e_{i}, d_{*}) = \log_{2}(\frac{  n_{d_{*}e_{i}}(0, 30) + 1/2   }{  
E^{*}_{d_{*}e_{i}}(0, 30) +1/2   
})
\end{equation}  
where, 
\begin{equation*}
E^{*}_{d_{*},e_{i}}(0, 30) =  \frac{ n_{d_{*} e_{i}}(-810, -630)  E_{d_{*} e_{i}}(0, 30) }{   E_{d_{*} e_{i}}(-810, -630)  }
\end{equation*}
When investigating possible ADRs for drug $d_{*}$ the authors use a filter to remove any Read Codes $e_{i}$ if the $IC(-30,-1, e_{i}, d_{*})> IC(0,30, e_{i}, d_{*})$ or $IC(0,0, e_{i}, d_{*})> IC(0,30, e_{i}, d_{*})$, these correspond to the $IC$ value the month prior to the prescription being higher than the month after prescription, or the $IC$ value on the day of prescription being higher than a month after.  In this paper we will use the $IC_{\Delta}(e_{i}, d_{*})$ as described above as an attribute for each possible drug and Read Code combination and use the Heaviside step function $H:\mathbb{R} \to \{ 0,1\}$, 
\begin{equation*}
\label{heavy}
H[n] = 
\begin{cases}
0 & \mbox{if } n \leq 0 \\
1 & \mbox{if } n > 0 \\
\end{cases}
\end{equation*}
to define the filter functions, 
\begin{equation}
\label{eq:filt}
\begin{aligned}
\zeta_{1}(e_{i}, d_{*})=&H[ IC(-30,-1, e_{i}, d_{*})- IC(0,30, e_{i}, d_{*}) ]\\ 
\zeta_{2}(e_{i}, d_{*})=&H[ IC(0,0, e_{i}, d_{*})- IC(0,30, e_{i}, d_{*})]\\
\end{aligned}
\end{equation}
as two addition binary attributes.

The algorithm Mining Unexpected Temporary Association Rules given the Antecedent (MUTARA) \cite{Jin2010} applies a case control approach that estimates the background rate that the Read Code is recorded into the THIN database by finding out how many patients who have not been prescribed the drug of interest have the Read Code recorded during a random time interval.  MUTARA aims to find Unexpected Temporary Association Rules (UTARs) by investigating how many patients have a specific Read Code unexpectedly recorded within a period of interest centred on the first prescriptions of the drug being studied.  The algorithm Highlighting UTARs Negating TARs (HUNT) \cite{Ji2012} was developed by the same authors as MUTARA and applies a similar method but is less prone to ranking therapeutic failure Read Codes (Read Codes linked to the cause of taking the drug) above Read Codes corresponding to ADRs.

MUTARA and HUNT both investigate the Read Codes that occur during the month after the drug being studied is first prescribed to a patient or the union of the month after the first and second prescriptions of the drug if it is repeated within a month of the first prescription. Previously we defined $\alpha(\omega, d_{*})$ to be the ages in days that patient $\omega$ is prescribed the drug $d_{*}$ and $\alpha_{1}(\omega, d_{*}))$ to be the age in days that patient $\omega$ is first prescribed the drug $d_{*}$, so the set $\alpha(\omega, d_{*}) \setminus  \alpha_{1}(\omega, d_{*})$ contains the ages that patient $\omega$ is prescribed drug $d_{*}$ except the age when they are first prescribe the drug.  Using this we define a new function that returns the patient's age in days for the second time a drug is prescribed for a patient if this is within a month of the first prescription or the patient's age in days for the first time the drug is prescribed if the second prescription is not within a month of the first.
\begin{equation}
\label{mut1}
\alpha_{2}(\omega, d_{*}) = 
\begin{cases}
\min \{s \in \alpha(\omega, d_{*}) \setminus \alpha_{1}(\omega, d_{*})\}  &   \mbox{if } \exists t \in \alpha(\omega, d_{*}) \setminus  \alpha_{1}(\omega, d_{*}) \mbox{ s.t } |t-\alpha_{1}(\omega, d_{*})|<30 \\
\alpha_{1}(\omega, d_{*}) &  \mbox{else} \\
\end{cases} 
\end{equation}

MUTARA uses patients that have not been prescribed the drug (so $\alpha_{1}(\omega, d_{*})=-1$) to estimate the background rate that the Read Code is prescribed into the THIN database.  For each patient that has never been prescribed the drug $d_{*}$, a random time interval is chosen within the age that the patient first has any Read Code recorded and the age that a Read Code is last recorded, $\min(A_{M}(\omega))$ and $\max(A_{M}(\omega))$ respectively, where we defined $A_{M}(\omega)$ previously to be the set of ages that patient $\omega$ has a Read Code recorded into the THIN database.  If a patient has never had a Read Code recorded into the THIN database ($A_{M}(\omega)=\emptyset$) or only had Read Codes recorded for the short period of time ($\max(A_{M}(\omega))-\min(A_{M}(\omega))<|t_{2}-t_{1}|$) then the patient is not used in this study as they are not an active patient and may bias results.  The time interval of length $|t_{2}-t_{1}|$ is chosen uniformly within [$\min(A_{M}(\omega))$...$\max(A_{M}(\omega))$].

Putting this all together, the time interval of interest around the first prescription of drug $d_{*}$ for patients prescribed the drug or the time interval chosen at random for patients who have never been prescribed the drug used by MUTARA is,
\begin{equation}
T_{M}(\omega, d_{*}, t_{1}, t_{2}) = 
\begin{cases}
[(\alpha_{1}(\omega, d_{*})+t_{1}) ... (\alpha_{2}(\omega, d_{*}) +t_{2})] & \mbox{if $\alpha_{1}(\omega, d_{*}) \neq -1$}\\
[ (\min(A_{M}(\omega))+r)... (\min(A_{M}(\omega))+r+|t_{2}-t_{1}|)]& \mbox{else}
\end{cases}
\end{equation}
where,  
\begin{equation}
r \sim U(0, \max(A_{M}(\omega))-\min(A_{M}(\omega))-|t_{2}-t_{1}|)
\end{equation}

For patients who are prescribed the drug $d_{*}$ a filter is applied to ignore any `expected' Read Codes that are recorded within the time interval of interest after the first prescription of the drug.  A Read Code is `expected' for the patient during the time interval of interest after the drug is prescribe if the patient also had the Read Code recorded within $t_{3} \in \mathbb{N}$ days prior to drug being prescribed.  We define this time interval prior to the prescription as,
\begin{equation}
T_{filt}(\omega, d_{*}, t_{3}) = 
\begin{cases}
[(\alpha_{1}(\omega, d_{*})-t_{3}) ... (\alpha_{1}(\omega, d_{*}) -1)] & \mbox{if $\alpha_{1}(\omega, d_{*}) \neq -1$}\\
\emptyset & \mbox{else}\\
\end{cases}
\end{equation}
 
We then find how many patients have each Read Code recorded `unexpectedly' during the time interval of interest after the drug by,
\begin{equation}
h^{*}(e_{i}, d_{*}, \omega, t_{1}, t_{2}, t_{3}) = \mathbbm{1} \{  e_{i} \in \underset{t \in T_{M}(\omega,d_{*}, t_{1}, t_{2}) }{\cup f_{M}(t, \omega)} \} 
(1-\mathbbm{1} \{ e_{i} \in \underset{t \in T_{filt}(\omega,d_{*}, t_{3}) }{\cup f_{M}(t, \omega)}  \})
\end{equation}
The Heaviside step function applied to $\alpha_{1}(\omega, d_{*})$ returns $1$ if patient $\omega$ has been prescribed the drug $d_{*}$, as in that case $\alpha_{1}(\omega, d_{*}) > 0$ or returns $0$ if patient $\omega$ has never been prescribed the drug $d_{*}$. MUTARA then calculates the unexpected-leverage as,
\begin{equation}
uL(e_{i},d_{*})  = \frac{ \sum_{\omega \in \Omega}{ [h^{*}(e_{i}, d_{*}, \omega, t_{1}, t_{2}, t_{3}) \times H[   \alpha_{1}(\omega, d_{*})] ]} }{|\Omega|}
             - \Big(\frac{\sum_{\omega \in \Omega}{H[ \alpha_{1}(\omega, d_{*})] }}{|\Omega|}\Big) \Big(\frac{\sum_{\omega \in \Omega}{ h^{*}(e_{i}, d_{*}, \omega, t_{1}, t_{2}, t_{3})}} {|\Omega|}\Big)
\end{equation} 
where the fraction of patients in the database who have the drug multiplied by the fraction of patients who have the Read Code recorded during their interval of interest is subtracted from the fraction of patients in the database who have the Read Code recorded within a month of the first time the drug is prescribed (or up to a month after the second prescription if the drug is repeated within a month).

HUNT calculates a similar value to the unexpected-leverage known as the leverage that does not include a filter to remove 'expected' Read codes based on a patient's history,
\begin{equation}
h^{**}(e_{i}, d_{*}, \omega, t_{1}, t_{2}) = \mathbbm{1} \{  e_{i} \in \underset{t \in T_{M}(\omega,d_{*}, t_{1}, t_{2}) }{\cup} f_{M}(t, \omega)  \}
\end{equation}
and
\begin{equation}
L(e_{i},d_{*})  = \frac{\sum_{\omega \in \Omega}{ [h^{**}(e_{i}, d_{*}, \omega, , t_{1}, t_{2}) \times H[  \alpha_{1}(\omega, d_{*})] ]} }{|\Omega|} - \Big(\frac{\sum_{\omega \in \Omega}{H[ \alpha_{1}(\omega, d_{*})] }}{|\Omega|}\Big) \Big(\frac{\sum_{\omega \in \Omega}{ h^{**}(e_{i}, d_{*}, \omega, , t_{1}, t_{2})}}{|\Omega|}\Big) 
\end{equation} 
HUNT then ranks the Read Codes in descending order of the leverage to attain the leverage rank and in descending order of the unexpected-leverage to attain the unexpected-leverage rank and then calculates the rank ratio between the leverage rank and unexpected-leverage rank.  Finally, HUNT returns the Read codes in descending order of this rank ratio. 

\subsection{Novel Algorithm Attributes}
The attributes of interest for detecting ADRs are how many patients have the Read Code recorded a month prior to the first prescription compared with how many patients have it recorded a month after,   
\begin{equation}
\label{eq:ab30}
ABratio_{30}(e_{i}, d_{*}) = \frac{\sum_{\omega \in \Omega}{h(e_{i}, d_{*}, \omega, 1, 30)}}{\sum_{\omega \in \Omega}{h(e_{i}, d_{*}, \omega, -30, -1)}}
\end{equation}   

a similar attribute considers the ratio between the number of patients who have the Read Code recorded a year after compared to a year before,
\begin{equation}
ABratio_{365}(e_{i}, d_{*}) = \frac{\sum_{\omega \in \Omega}{h(e_{i}, d_{*}, \omega, 1, 365)}}{\sum_{\omega \in \Omega}{h(e_{i}, d_{*}, \omega, -365, -1)}}
\end{equation} 

As the THIN database does not record the time that a Read Code or drugcode is recorded, Read Codes that are recorded on the same day as the drugcode could correspond to a serious ADR that occurs immediately or an indicator, this is why the day of prescription is not included in the above $ABratios$ and instead is used as another attribute,
\begin{equation}
DOP(e_{i}, d_{*}) = \frac{\sum_{\omega \in \Omega}{h(e_{i}, d_{*}, \omega, 0, 0)}}{\sum_{\omega \in \Omega}{h(e_{i}, d_{*}, \omega, -365, -1)}}
\end{equation}   

In previous methods there has been a patient level filter that removes medical events from a patient's sequence that occur $30$ days after the drug if the patient also had the medical event shortly prior to the drug \cite{Jin2006}. The justification for this is, if a patient had an illness shortly before the drug then having the illness repeated after the drug is unlikely to be an ADR. This inspires the third attribute that is based on the number of patients who have the Read Code recorded in the month after and also have it recorded during the month prior to the first prescription,
\begin{equation}
\label{eq:expect}
expect(e_{i}, d_{*}) = \frac{\sum_{\omega \in \Omega}{[h(e_{i}, d_{*}, \omega, 1, 30) \times h(e_{i}, d_{*}, \omega, -30, -1) ]}}{\sum_{\omega \in \Omega}{h(e_{i}, d_{*}, \omega, 1, 30)}}
\end{equation}  

The final attributes of interest make use of the Read Code tree structure.  When a patient first has an illness it is likely that not much detail is known, so a low level Read Code will probably be recorded into the THIN database, over time more information may be discovered about a patient's illness possibly due to laboratory results and this may then result in a more specified higher level Read Code being entered (a child of the original less detailed Read Code).  As a consequence, it is common for higher level Read Codes related to the cause of taking the drug to only be recorded after the drug, so to help distinguish between these and ADRs we calculate the Read Code after and before ratio when only considering the first two or three elements of a Read Code. For example, considering the first three elements of the Read Code, the Read Code $A11ab$ becomes $A11$ and the $ABratio_{lev3}$ is the number of patients that have a Read Code starting with $A11$ within the month after the first prescription of the drug divided by the number of patients that have a Read Code starting with $A11$ within the month prior to the first prescription of the drug.  Similarly, the $ABratio_{lev2}$ for a Read Code is the number of patients who have a Read Code with the same first two elements within a month after the first prescription divided by the number of patients who have a Read Code with the same first two elements within a month prior to the prescription.

To calculate these attributes we first define an equivalence relationship, $e_{i} \overset{n}{\sim} e_{j}$ if the first $n$ elements of the Read Codes of $e_{i}$ and $e_{j}$ are the same.  The number of patients who have Read Codes equivalent to $e_{i}$ recorded a month after the first prescription of $d_{*}$ is,
\begin{equation*}
\sum_{\omega \in \Omega}{ H[\sum_{e \overset{n}{\sim} e_{i}}{h(e, d_{*}, \omega, 1, 30)}] }
\end{equation*}
similary, the number of patients who have Read Codes equivalent to $e_{i}$ recorded during the month prior to the prescription is,
\begin{equation*}
\sum_{\omega \in \Omega}{ H[\sum_{e \overset{n}{\sim} e_{i}}{h(e, d_{*}, \omega, -30, -1)}] }
\end{equation*}
It follows that the $ABratio_{lev3}$ and $ABratio_{lev2}$ are calculated by,
\begin{equation}
ABratio_{lev3}(e_{i}, d_{*}) = \frac{\sum_{\omega \in \Omega}{ H[\sum_{e \overset{3}{\sim} e_{i}}{h(e, d_{*}, \omega, 1, 30)}] } }{\sum_{\omega \in \Omega}{ H[\sum_{e \overset{3}{\sim} e_{i}}{h(e, d_{*}, \omega, -30, -1)}] }}
\end{equation} 
\begin{equation}
\label{ablev3}
ABratio_{lev3}(e_{i}, d_{*}) = \frac{\sum_{\omega \in \Omega}{ H[\sum_{e \overset{2}{\sim} e_{i}}{h(e, d_{*}, \omega, 1, 30)}] } }{\sum_{\omega \in \Omega}{ H[\sum_{e \overset{2}{\sim} e_{i}}{h(e, d_{*}, \omega, -30, -1)}] }}
\end{equation} 

\section{DRESS Algorithm}
\label{sec:dress}
\begin{figure}[!t]
\centering
\includegraphics[trim=40 0 0 20,clip,width=0.45\textwidth]{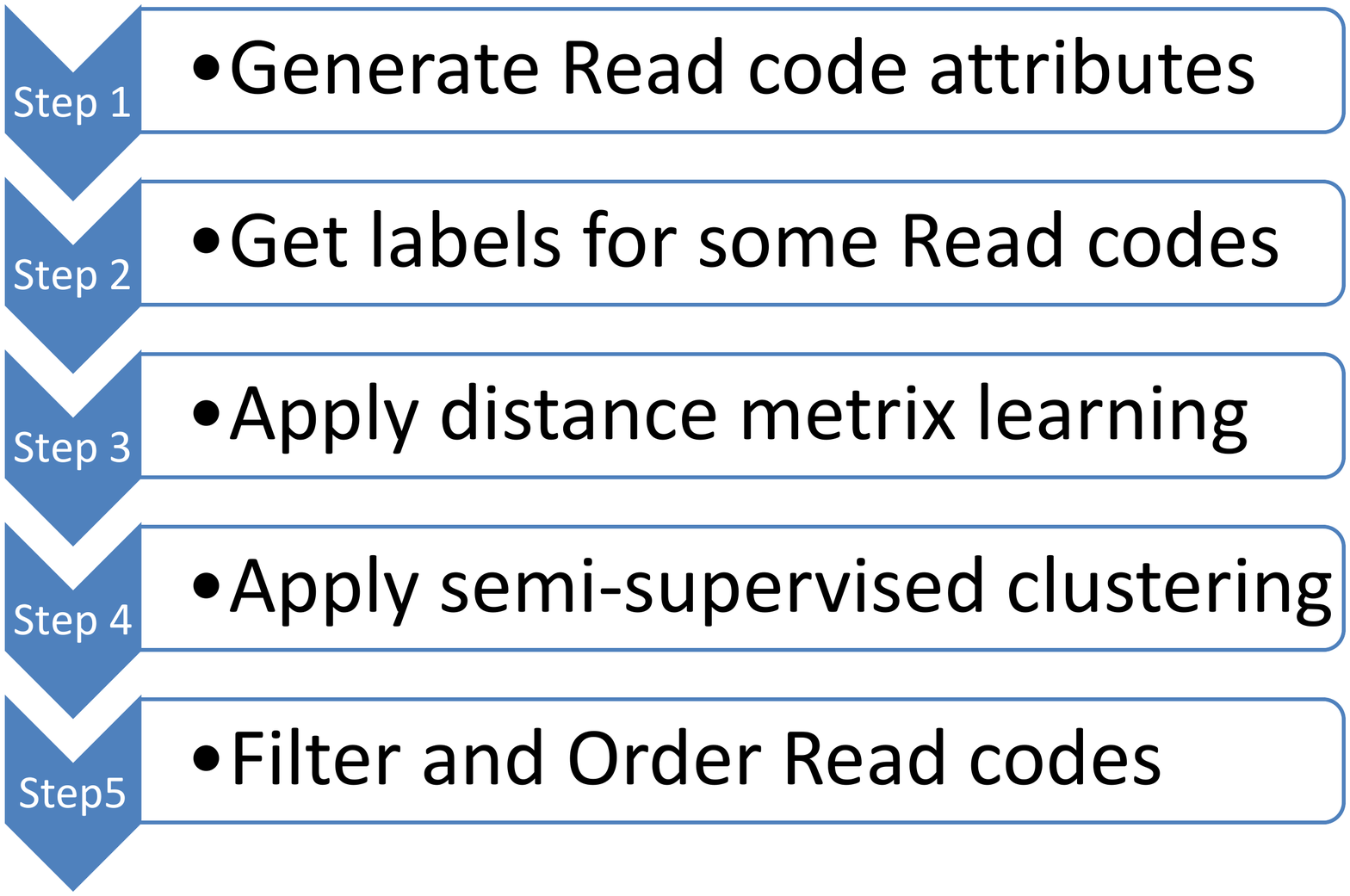}
\caption{A summary of the five steps applied in the DRESS algorithm.}
\label{fig:method}
\end{figure}
The Detecting Rare Events Semi-Supervised (DRESS) algorithm comprises of five steps, see Fig (\ref{fig:method}).  The DRESS algorithm requires the user to input the drug of interest ($d_{*}$) and returns a ranked list of Read Codes in descending order of how likely they are to be ADRs.  Tentative ADR signals can then be determined by the DRESS algorithm by considering the top $k$ ranked Read codes to be signalled as ADRs.  The value of $k$ will determine the filtering threshold and in this paper we consider the top 100 ranked Read Codes to be signalled by the algorithms. 

\subsection{Step 1}
The first step of the DRESS algorithm is the generation of attributes for any Read Code that could be an ADR. This is accomplished by initially finding the set of all the Read Codes that are recorded within a month of the first prescription of $d_{*}$ for any patient, we denote this set of Read Codes by $G$. Then, for every Read Code $e_{i} \in G$ the DRESS algorithm generates the attributes $ABratio_{30}(e_{i},d_{*})$, $ABratio_{365}(e_{i},d_{*})$, $DOP(e_{i},d_{*})$, $expect(e_{i},d_{*})$, $ABratio_{lev3}(e_{i},d_{*})$, $ABratio_{lev2}(e_{i},d_{*})$ (Eq's (\ref{eq:ab30})-(\ref{ablev3})) and the OE ratio values the $IC_{\Delta}(e_{i},d_{*})$, $\zeta_{1}(e_{i},d_{*})$ and $\zeta_{2}(e_{i},d_{*})$ (Eq's (\ref{eq:icdelta})-(\ref{eq:filt})).  The data point for each $e_{i} \in G$ is denoted by $\mathbf{x_{i}} \in \mathbb{R}^{9}$ and is a vector corresponding to the attribute values.  The set of data points is denoted by $X=\{\mathbf{x_{1}},\mathbf{x_{2}},...,\mathbf{x_{n}}  \}$, where the cardinality of the set $G$ is $n$ ($|G|=n$).   

\subsection{Step 2}
\begin{figure*}[t]
\centering
\includegraphics[trim=20 80 20 40,clip,width=0.95\textwidth]{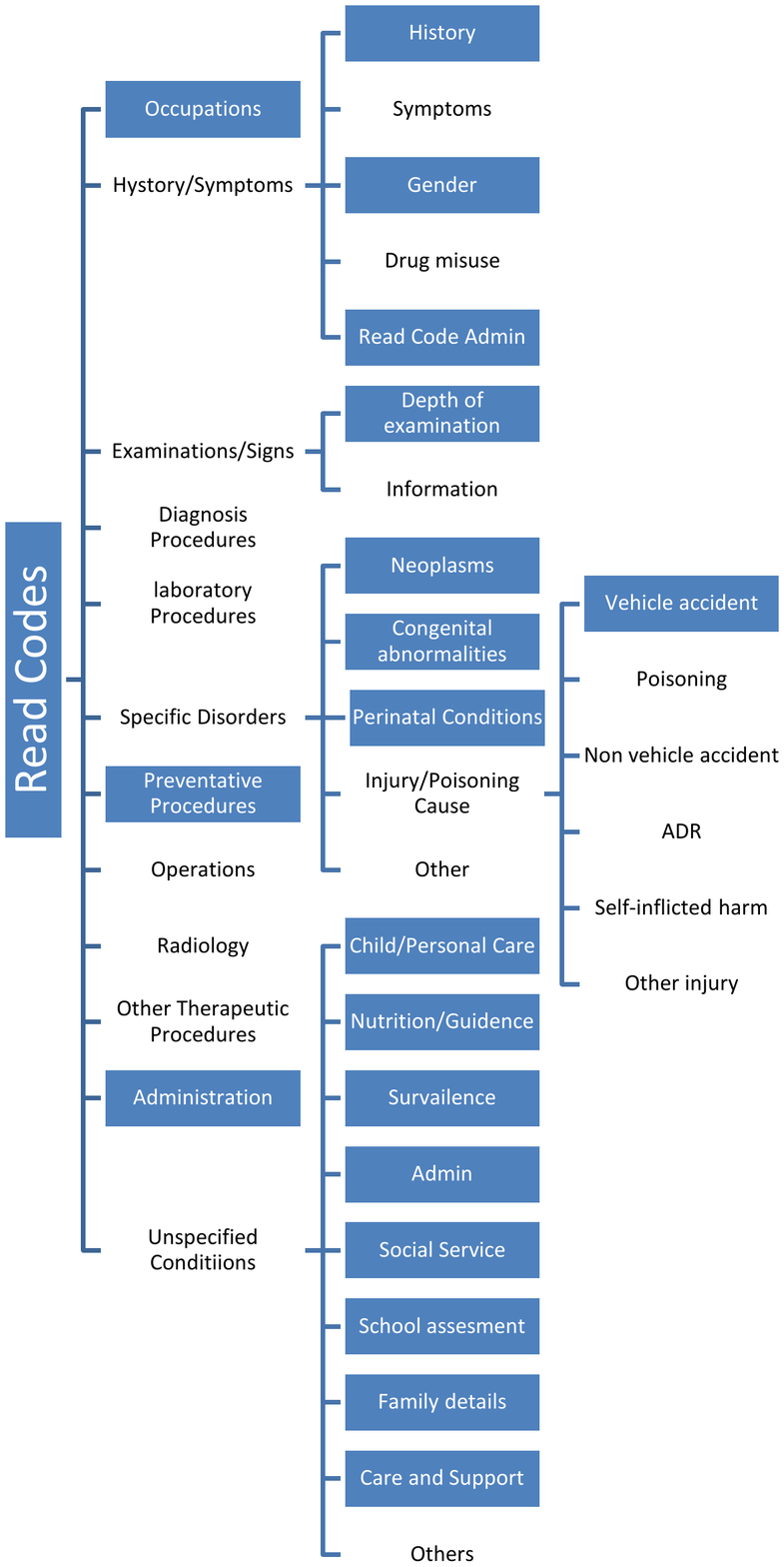}
\caption{The Read Codes categories labelled as noise are in the shaded boxes. Read Codes for non-medical events such as occupation, vehicle accidents and administration do not offer information about immediately occurring ADRs as well as congenital/perinatal or cancer events so we label these as noise.}
\label{fig:ignore}
\end{figure*}
As we are applying a semi-supervised approach, we need labels for some of the Read Codes.  We have decided to have three different labels for each Read Code, one label representing Read Codes that are ADRs (known ADR), another label representing Read Codes that cause the drug to be taken by the patients (indicator) and the final label representing Read Codes that are not linked to the drug but just occur by chance (noise).  The reason for choosing three labels is because there is information available to enable us to determine the labels for a sufficient number of Read Codes when using three labels.  

We determined the Read Codes that are labelled as noise by using the hierarchal structure of the Read Code tree. We determined branches that are not related to immediately occurring ADRs by manually investigating the Read Code tree and found the set of irrelevant Read Codes, $M_{irrel}$. Examples of Read Codes in $M_{irrel}$ that cannot correspond to immediately occurring ADRs are those related to cancer, occupations or family history.  Fig (\ref{fig:ignore}) illustrates the Read Code tree and the Read Code branches considered to be noise are shaded in, whereas the Read Code branches that are possible ADRs are unshaded.  Using the set $M_{irrel}$, the DRESS algorithm labels any Read Code $e_{i} \in G\cap M_{irrel}$ as noise.

To determine the labels of some of the Read Codes corresponding to known ADRs or indicators we mined data from the internet.  The Read Codes labelled as indicators are found by first extracting the strings listed as indicators on the netdoctor website \cite{netdr2012}, then finding the set of Read Codes with descriptions containing any one of these strings and finally validating these are indicators by ignoring indicator Read Codes that do not have an $ABratio_{30}<1$.  The Read Codes labelled as known ADRs are found similarly, by first extracting the strings listed as side effects on the netdoctor website, then finding the set of Read Codes with descriptions containing any one of these strings and validating these by ignoring any Read Codes that do not have an $ABratio_{30} \geq 1.5$.  In addition to the Read Codes corresponding to the netdoctor listed side effects, Read Codes with a description containing the drug name and the term `adverse' are also labelled as known ADRs.

\subsection{Step 3}
After labelling some of the Read Codes we then applied the metric learning algorithm detailed in \cite{Ying2012}.  Letting $S$ denote the set of all index pairs for Read Codes with the same label (e.g. if Read Codes corresponding to the data points $\mathbf{x_{1}}$ and $\mathbf{x_{2}}$ are both labelled as noise then $(1,2)\in S$), $D$ denote the set of all index pairs for Read Codes with a different label (e.g. if the Read code corresponding to data point $\mathbf{x_{1}}$ is labelled as noise but the Read code corresponding to the data point $\mathbf{x_{5}}$ is labelled as an indicator then $(1,5) \in D$) and the inner product of two $d \times n$ real valued matrices, $X,Y \in \mathbb{R}^{d \times n}$, is denoted by $\langle X,Y\rangle :=$\textbf{Tr}$(X^{T}Y)$, where \textbf{Tr}(A) means the trace of the matrix $A$.  The cone of positive semidefinite matrices is denoted by $S_{+}^{d}$.  

Given a pair of Read Code data points $\mathbf{x_{i}}$ and $\mathbf{x_{j}}$ generated in step 1, we calculate the matrix $X_{ij}=(\mathbf{x_{i}}-\mathbf{x_{j}})(\mathbf{x_{i}}-\mathbf{x_{j}})^{T}$. If $\tau=(i,j)$ is an index pair, then $X_{\tau}\equiv X_{ij}$.  The matrix $X_{S}$ is defined by $X_{S}=\sum_{(i,j) \in S }{X_{ij}}$ and $\overset{\sim}{X}_{\tau} = X_{S}^{-1/2} X_{\tau} X_{S}^{-1/2}$. The authors calculated that $\nabla f_{\mu}(S_{t}^{\mu})= 
\frac{ \sum_{\tau \in D}{e^{-\langle \overset{\sim}{X}_{\tau}, S \rangle/\mu}\overset{\sim}{X}_{\tau}}  }{   \sum_{\tau \in D}{e^{-\langle \overset{\sim}{X}_{\tau}, S \rangle}/\mu} }$.

\begin{algorithm}
 \SetKwInOut{Input}{Input}\SetKwInOut{Output}{Output}
 \SetAlgoLined
 \Input{
 \begin{itemize}
 \item smoothing parameter $\mu >0$ (e.g., $10^{-5}$)\\
 \item tolerance value $tol$ (e.g., $10^{-5}$) \\
 \item step sized $\{\alpha_{t} \in (0,1):t \in \mathbb{N} \}$
 \end{itemize}
 }
 \Output{$d \times d$ matrix $S_{t}^{\mu} \in S_{+}^{d}$ }
 \textbf{Initialization}: $S_{1}^{\mu} \in S_{+}^{d}$ with \textbf{Tr}($S_{1}^{\mu}$)$=1$ \\
 \For{$t=1,2,3,...$}{
$Z_{t}^{\mu}$=argmax$\{f_{\mu}(S_{t}^{\mu}) + \langle Z, \nabla f_{\mu}(S_{t}^{\mu}) \rangle  : Z \in S_{+}^{d}$, \textbf{Tr}(Z)=1  $\}$, that is, $Z_{t}^{\mu}=\boldsymbol{\nu}\boldsymbol{\nu}^{T}$ where $\boldsymbol{\nu}$ is the maximal eigenvector of the matrix $\nabla f_{\mu}(S_{t}^{\mu})$ \\
$S_{t+1}^{\mu} = (1-\alpha_{t})S_{t}^{\mu}+\alpha_{t}Z_{t}^{\mu}$\\
if $|f_{\mu}(S_{t+1}^{\mu}) - f_{\mu}(S_{t}^{\mu})| < tol$ then \textbf{break}
 }
 \caption{The distance metric learning algorithm from \cite{Ying2012}}
 \label{alg:metric}
\end{algorithm}

When applied in the DRESS algorithm, the distance metric algorithm described in algorithm \ref{alg:metric} finds a mapping from the space consisting of the attribute values found in step 1 to a different space that minimises the distance between two Read Codes with the same label and maximises the distance between two Read Codes with different labels.  The mapping is, $f:\mathbb{R}^{9} \to \mathbb{R}^{9}; f(\mathbf{x_{i}})=\mathbf{x_{i}}^{T}S_{t}^{\mu} $, where $S_{t}^{\mu} $ is the $9 \times 9$ learned distance metric matrix,

\subsection{Step 4}
The constrained K-means algorithm \cite{Basu2002} is applied to the Read Codes on their corresponding transformed data points determined by the distance metric learning algorithm above.  The constrained K-means algorithm is a semi-supervised algorithm that fixes the class of the labelled Read Codes and uses these labelled Read Codes to calculate the initial cluster centres then iteratively assigns the non-fixed Read Codes into the cluster with the closest mean, with the means iteratively being recalculated until convergence.  

When the DRESS algorithm is implemented, the set of data points input into the Constrained K-means algorithm is the set $\{ f(\mathbf{x_{1}}), f(\mathbf{x_{2}}), ..., f(\mathbf{x_{n}})  \}$, the value of $K$ input is $3$ and the initial seeds are $S_{1}=\{\mathbf{x_{i}}: \mathbf{x_{i}} \mbox{ is labelled as a known ADR}\}$, $S_{2}=\{\mathbf{x_{i}}: \mathbf{x_{i}} \mbox{ is labelled as an indicator}\}$ and $S_{3}=\{\mathbf{x_{i}}: \mathbf{x_{i}} \mbox{ is labelled as noise}\}$.  

\begin{algorithm}[H]
 \SetKwInOut{Input}{Input}\SetKwInOut{Output}{Output}
 \SetAlgoLined
 \Input{
 Set of data points $X=\{\mathbf{x_{1}}, \mathbf{x_{2}},...,\mathbf{x_{n}} \}$, $\mathbf{x_{i}} \in \mathbb{R}^{d}$, number of clusters $K$, the set $S=\cup_{l=1}^{K}S_{l}$ of initial seeds.
 }
 \Output{
 Disjoint $K$ partitioning $\{ X_{l}\}_{l=1}^{k}$ of $X$ such that the KMeans objective function is optimised.
 }
 \textbf{Initialization}: $\boldsymbol{\mu}_{h}^{(0)} \leftarrow \frac{1}{|S_{h}|}\sum_{\mathbf{x} \in S_{h}}{\mathbf{x}}$, for $h=1,...,K$; $t \leftarrow 0$ \\
 \Repeat{convergence}{
 For $\mathbf{x} \in S$, if $\mathbf{x} \in S_{h}$ assign $\mathbf{x}$ to the cluster $h$ (i.e., set $X_{h}^{t+1}$).  For $\mathbf{x} \not\in S$, assign $\mathbf{x}$ to the cluster $h^{*}$ (i.e., set $X_{h^{*}}^{t+1}$), for $h^{*}=\underset{h}{\mbox{argmin}}\|\mathbf{x}-\boldsymbol{\mu}_{h}^{(t)}\|^{2}$ \\
 $\boldsymbol{\mu}_{h}^{(t+1)} \leftarrow \frac{1}{|X_{h}^{(t+1)}|} \sum_{\mathbf{x} \in X_{h}^{(t+1)}} x$\\
 $t \leftarrow (t+1)$\\ 
 }
 \caption{The Constrained K-means algorithm developed in \cite{Basu2002}}
  \label{alg:kmeans}
\end{algorithm}

Read Codes in the same cluster as the Read Codes that were originally labelled as known ADRs are referred to as being in the ADR cluster, Read Codes in the same cluster as the Read Codes that were originally labelled as indicators are referred to as being in the indicator cluster and Read Codes in the same cluster as the Read Codes that were originally labelled as noise are referred to as being in the noise cluster. 

\subsection{Step 5}
The last step involved applying two additional filters and then used the Read Code attributes and clustering to order the Read Codes by how likely they are to be ADRs.  The first filter removed all the Read Codes ($e_{i}$'s) that were in the indicator cluster or where $(1-expect(e_{i}, d_{*})) \times ABratio_{30}(e_{i}, d_{*})<1$ .  The second filter is a filter that we have developed for post processing with any algorithm that detects ADRs by mining the THIN database.  This filter removes all the Read Codes that are irrelevant for ADR detection such as Read Codes corresponding to administrative events.  Finally Read Codes were ordered in descending order of $(1-expect(e_{i}, d_{*})) \times ABratio_{30}(e_{i}, d_{*}) \times \frac{1}{\beta}$, where $\beta= 1$ for Read Codes in the ADR cluster and $\beta=3$ for Read Codes in the noise cluster.

In summary, the DRESS algorithm uses the semi-supervised clustering to filter Read Codes that have attributes that make them unlikely to be ADRs and then orders the remaining Read Codes by how often they occurred unexpectedly after the first prescription of the drug being investigated but also adds a weight so that Read Codes that have attributes similar to known ADRs are ranked higher.   

\section{Results}
\label{sec:results}
\begin{sidewaystable}[tp]
\centering

\caption{The Read Code ranks that serious and rare ADRs are given by the different methods for a range of drug and known rare ADR pairs. If the algorithm did not return a rank for the Read Code corresponding to the ADR, then this is represented by `-'.}
\label{tab:adr}
\scalebox{0.85}{
\begin{tabular}{ccccc|cccc}
Drug & Indication & ADR & Withdrawn & Incidence & DRESS & OE ratio & HUNT & MUTARA \\
\hline
Rofecoxib & Acute pain & Inferior myocardial infarction & 2004 & 0.79-1.24\% \cite{Graham2005} & 83 & 302 & 196 & 1212 \\
Rofecoxib & Acute pain & Heart failure & 2004 & 0.79-1.24\% \cite{Graham2005} & 121 & 805 & 504 & 562 \\
Rimonabant & Obesity & Depressed mood & 2008 & 0.6\% \cite{Despres2005} & 16 & 32 & 15 & 68 \\
Rimonabant & Obesity & Depressive disorder NEC &2008 & 3 \% \cite{Christensen2007}& 23 & 180 & 295 & 1157 \\ 
Enalapril & Hypertension/Heart failure& Acute pancreatitis & - & rare \cite{Carnovale2003} \cite{Balani2008} & 53 & 700 & 416 & 2081 \\ 
Naproxen & Pain/inflammation & Acute renal failure & - & rare \cite{Harirforoosh2009} & 20 & 159 & 138 & 1313 \\
Naproxen & Pain/inflammation & Jaundice & - & rare \cite{Ali2011} & 52 & 774 & 3394 & 4190 \\
Naproxen & Pain/inflammation & Hepatitis unspecified & - & rare \cite{Ali2011} & 241 & 35 & 852 & 3502 \\
Naproxen & Pain/inflammation & Suicide and selfinflicted injury & - & rare \cite{Onder2004} & 104 & 223 & 242 & 1808 \\

Nifedipine & Hypertension/Angina & Weight increasing & - & $\leq 1$\% \cite{drugs2012} & 362 & 29 & 1517 & 1492 \\
Nifedipine & Hypertension/Angina  & Feels hot/feverish & - & $\leq 2$\% \cite{drugs2012} & 314 & 428 & 1972 & 7480 \\

Ciprofloxacin & Bacterial infections & Nontraumatic rupture of Achilles tendon & - & rare \cite{Linden 2002} & 13 & 42 & 1273 & 5506\\
Levofloxacin & Bacterial infections & Rupture Achilles tendon& - & rare \cite{Linden 2002} & 76 & 296 & - & -  \\
Levofloxacin & Bacterial infections & Nontraumatic rupture of Achilles tendon & - & rare \cite{Linden 2002} & 298 & 279 & 320 & 1189  \\

Rifampicin & Tuberculosis/infections & Thrombocytopenia NOS & - & rare \cite{Lee1989} & 75 & 418 & 265 & 1003 \\
Metformin & Type 2 diabetes & Lactic acidosis & - & 0.005 \%  \cite{Chan1999} & - & - & - & - \\
Metformin & Type 2 diabetes & Acidosis & - & 0.005 \%  \cite{Chan1999} & 449 & 813 & 466 & 3223 \\
\end{tabular}
}
\end{sidewaystable}

We applied the DRESS algorithm to a range of drugs, some of which have been withdrawn from the market due to serious ADRs.  Table \ref{tab:adr} shows the rank that each data mining algorithm assigns for a known rare and serious drug and ADR pair.  The table also states the cause for taking the drug (the indication), the year the drug was withdrawn and approximately how commonly the ADR occurs for patients prescribed the drug.  In some cases the rare and serious ADR being investigated was listed as an ADR on the netdoctor website and to prevent any bias in the results for the DRESS algorithm, any labels for the Read Codes corresponding to the ADR being investigated were removed at the end of step 2, prior to the semi-supervised steps.  So the Read Codes corresponding to the ADR being investigated were always unlabelled in the DRESS algorithm.

The DRESS algorithm had an average rank over the drug and ADR pairs of 143.75 and had the highest rank compared to all the other algorithms for 12 of the 16 drug and ADR pairs investigated.  It was able to return a rank under 100 for 56.25\% of the ADRs and all the ADRs had a rank below 500.  In comparison the other methods had an average rank of 344.69, 791 and 2385.73 for the OE ratio, HUNT and MUTARA respectively.  The OE ratio only returned a rank under 100 for 25\% of the ADRs and only 6.25\% of ADRs ranked by MUTARA and HUNT had a rank under 100.  

\section{Discussion}
\label{sec:disc}
The existing methods for detecting ADRs using EHDs do not currently definitively detect ADRs, but rather, they can be considered to generate tentative ADR signals for the top $k$ ranked medical events in their returned list.  This is effectively filtering out all the medical events that are unlikely to be ADRs or the medical events without sufficient evidence (number of patients experiencing the event after the drug) of being an ADR.  The medical events with a rank greater than $k$ are ignored and the medical events with a rank less than $k$ (the signalled medical events) are investigated further to confirm if they are true ADRs.  Therefore, for an unknown ADR to be detected by these existing methods it needs to be signalled by being ranked in the top $k$ medical events returned and the closer that the rank is to the value $1$, the more likely it will be investigated further, even when low values of $k$ are used.  The results of this paper show that the existing methods are not able to rank the known rare and serious ADRs highly, so they would be unlikely to signal these for further investigation, but the DRESS algorithm was able to rank over 50\% of these rare ADRs within the top 100 and would most likely signal these for further investigation.  This implies that the DRESS algorithm is more suitable than existing methods for detecting rare ADRs and has the potential to discover many unknown rare ADRs.

The DRESS algorithm was unable to rank `Naproxen and Hepatitis' and `Nifedipine and weight increase' higher than the OE ratio, one reason for this is that the DRESS algorithm does not perform as well for medical events that have a high background rate as medical events that are common will have a greater number of patients experiencing the medical event in the month before the prescription and if the ADR is rare only one or two patients extra will have the medical event in the month after the prescription, so the $ABratio$ will be close to one, but on the other hand, the OE ratio performs better for medical events with a high background rate as the bias in the $IC$ calculation will have less impact.  It might be better to have two ADR clusters, one for the medical events that occur at a low background rate and another for the medical events that occur at a high background rate, but this may cause issue if there is not a sufficient number of Read Codes corresponding to known side effects, as if there are only a few Read Codes for the known side effects there may only be one to two labelled Read Codes in each cluster and this will be deleterious for the semi-supervised methods.   

It may be argued that the DRESS algorithm can only be applied after many ADRs are known and this may prevent it efficiently detecting rare ADRs, but rare ADRs that occur less than 1 in 1000 patients generally need three of more ADR cases before there is satisfactory evidence to confirm the ADR and this requires thousands of patients having the drug.  DRESS can be implemented after a few hundred patients have taken a drug as the obvious side effects will be known, so the constraint of requiring known ADRs will not reduce the efficiency of DRESS.  It is worth noting that the current implementation of DRESS will not be as effective if a drug is generally safe and does not have many side effects, but this is not common and DRESS could be modified to use the positive effects of the drug as labelled medical events as the positive effects and side effects have similar attribute values as both should increase after the drug is taken.

The DRESS algorithm still highly ranks medical events related to the cause of the drug and removing these medical events would greatly improve the ability to detect rare ADRs with DRESS.  The $expect$ attribute has limitations as it only indicates if the patient had a repeat of a medical event and does not make use of the medical event relations within the THIN database to determine if a medical event is expected based on related medical events.  For example, if a patient has `a cold' then they are likely to experience `a cough' if the illness progresses, but the $expect$ attribute only says `a cough' is expected if the patient has had `a cough' shortly before and not if a predecessor medical event has occurred before.  To reduce the rank of medical events related to the cause of the drug a new attribute that uses sequential patterns to determine the expectedness of each medical event that occurs after the description could be used.

\section{Conclusion}
\label{sec:con}
In this paper we have described a novel methodology to detect rare ADRs that incorporates some existing methods (Observe Expected ratio, MUTARA and HUNT), information retrieval from the web, metric learning and semi-supervised clustering.  The results suggest this methodology is able to detect rare and serious ADRs for the range of drugs chosen in the investigation and has the potential to help detect many currently unknown ADRs.  The method is not able to remove all medical events related to the cause of taking the drug and future work should aim to prevent generating signals for these medical events by adding an additional attribute to the clustering that determines the expectedness of each medical event based on sequential patterns that can be mined from the whole database.  Further work could also investigate different metric learning and semi-supervised clustering techniques.


%





\ifCLASSOPTIONcaptionsoff
  \newpage
\fi



%
\bibliographystyle{IEEEtran}
\bibliography{dressed}

\begin{thebibliography}{10}
\providecommand{\url}[1]{#1}
\csname url@samestyle\endcsname
\providecommand{\newblock}{\relax}
\providecommand{\bibinfo}[2]{#2}
\providecommand{\BIBentrySTDinterwordspacing}{\spaceskip=0pt\relax}
\providecommand{\BIBentryALTinterwordstretchfactor}{4}
\providecommand{\BIBentryALTinterwordspacing}{\spaceskip=\fontdimen2\font plus
\BIBentryALTinterwordstretchfactor\fontdimen3\font minus
  \fontdimen4\font\relax}
\providecommand{\BIBforeignlanguage}[2]{{%
\expandafter\ifx\csname l@#1\endcsname\relax
\typeout{** WARNING: IEEEtran.bst: No hyphenation pattern has been}%
\typeout{** loaded for the language `#1'. Using the pattern for}%
\typeout{** the default language instead.}%
\else
\language=\csname l@#1\endcsname
\fi
#2}}
\providecommand{\BIBdecl}{\relax}
\BIBdecl

\bibitem{Gautier2003}
S.~Gautier, H.~Bachelet, R.~Bordet, and J.~Caron, ``The cost of adverse drug
  reactions,'' \emph{Exoert Opin Pharmacother}, vol. 4(3), pp. 319--26, 2003.

\bibitem{Patel2007}
K.~Patal, M.~Kedia, D.~Bajpai, S.~Mehta, N.~Kshirsagar, and N.~Gogtay,
  ``Evaluation of the prevalence and economic burden of adverse drug reactions
  presenting to the medical emergency department of a tertiary referral
  centre:a prospective study,'' \emph{{BMC} Clinical Pharmacology}, vol. 7:8,
  2007.

\bibitem{Davies2009}
E.~C. Davies, C.~F. Green, S.~Taylor, P.~R. Williamson, D.~R. Mottram, and
  M.~Pirmohamed, ``Adverse drug reactions in hostpitals in-patients: A
  prospectve analysis of 3695 patient-episodes,'' \emph{[PL]o[S] {ONE}}, vol.
  4(2), p. e4439.doi:10.1371/journal.pone.0004439, 2009.

\bibitem{Hartholt2010}
K.~A. Hartholt, N.~van~der Velde, C.~W.~N. Looman, M.~J.~M. Panneman, and E.~F.
  van Beeck, ``Adverse drug reactions related hostpital admissions in persons
  aged 60 years and over, thenetherlans, 1981-2007: Less rapid increase,
  different drugs,'' \emph{{PL}o{S} {ONE}}, vol. 5(11), p. e13977.
  doi:10.1371/journal.pone.0013977, 2010.

\bibitem{Shepherd2012}
G.~Shepherd, P.~Mohorn, K.~Yacoub, and D.~May, ``Adverse drug reaction deaths
  reported in united sates vital statistics, 1999-2006,'' \emph{Ann
  Pharmacother}, vol. 46(2), pp. 169--75, 2012.

\bibitem{Betteridge2012}
T.~M. Betteridge, C.~M. Frampton, and D.~L. Jardine, ``Polypharmacy - we make
  it worse! a cross-sectional study from an acute admissions unit,''
  \emph{Internal Medicine Journal}, vol. 42(2), pp. 208--211, 2012.

\bibitem{Pirmohamed2004}
M.~Pirmohamed, ``Adverse drug reactions as cause of admission to hospital:
  prospective analysis of 18 820 patients,'' \emph{BMJ}, vol. 329, pp. 15--19,
  2004.

\bibitem{Varallo2011}
F.~R. Varallo, M.~F.~R. Lima, J.~C.~F. Galduroz, and P.~C. Mastroianni,
  ``Adverse drug reactions as cause of hospital admission of elderly people: a
  pilot study,'' \emph{Latin American Journal of Pharmacy}, vol. 30(2), pp.
  347--353, 2011.

\bibitem{Berlin2008}
J.~A. Berlin, S.~C. Glasser, and S.~S. Ellenberg, ``Adverse event detection in
  drug development: Recommendations and obligations beyond phase 3,'' \emph{Am
  J Public Health}, vol. 98(8), pp. 1336--1371, 2008.

\bibitem{Seruga2011}
B.~Seruga, L.~Sterling, L.~Wang, and I.~F. Tannock, ``Reporting of serious
  adverse drug reactions of targeted anticancer agents in pivotal phass iii
  clinical trials,'' \emph{JCO}, vol. 29(2), pp. 174--185, 2011.

\bibitem{Berman2000}
L.~Bergman, M.~L. Beelen, M.~P. Gallee, H.~Hollema, J.~Benraadt, and F.~E. van
  Leeuwen, ``Risk and prognosis of endometrial cancer after tamoxifen from
  breath cancer. comprehensive cancer centers' {ALERT} group. assessment of
  liver and endometrical cancer risk following tamoxifen.'' \emph{Lancet}, vol.
  356(9233), pp. 881--887, 2000.

\bibitem{Ladewski2003}
L.~A. Ladewski, S.~M. Belknap, J.~R. Nebeker, O.~Sartor, E.~A. Lyons, T.~C.
  Kuzel, M.~S. Tallman, D.~W. Raisch, A.~R. Auerbach, G.~T. Schumock, H.~C.
  Kwaan, and C.~L. Bennett, ``Dissemination of information on potentially fatal
  adverse drug reactions for cancer drugs from 2000 to 2002: First results from
  the research on adverse drug events and reports project,'' \emph{J Clin
  Oncol}, vol.~21, pp. 3859--3866, 2003.

\bibitem{Evans2001}
S.~J.~W. Evans, P.~C. Waller, and S.~Davis, ``Use of proportional reporting
  ratios ({PRRs}) for signal generation from spontaneous adverse drug reaction
  reports,'' \emph{Pharmacoepidemiology and Drug Safety}, vol. 10(6), pp.
  483--486, 2001.

\bibitem{Bate1998}
A.~Bate, M.~Lindquist, I.~R. Edwards, S.~Olsson, R.~Orre, A.~Lansner, and
  R.~M.~D. Freitas, ``A bayesian neural network method for adverse drug
  reaction signal generation,'' \emph{European Journal of Clinical Pharmacol},
  vol.~54, pp. 315--321, 1998.

\bibitem{Dumouchel1999}
W.~DuMouchel, ``Bayesian data mining in large frequency tables, with an
  application to the fda spontaneous reporting system,'' \emph{American
  Statistician}, vol. 1999, pp. 177--190, 53.

\bibitem{Hazell2006}
L.~Hazell and S.~A.~W. Shakir, ``Under-reporting of adverse drug reactions: A
  systematic review,'' \emph{Drug Safety}, vol. 29(5), pp. 385--396, 2006.

\bibitem{Noren2010}
G.~M. Noren, J.~Hopstadius, A.~Bate, K.~Star, and I.~R. Edwards, ``Temportal
  pattern deiscovery in longitudinal electronic patients records,'' \emph{Data
  Mining and Knowledge Discovery}, vol. 2010, pp. 361--387, 20.

\bibitem{Schuemie2011}
M.~J. Schuemie, ``Methods for drug safety signal detection in longitudinal
  observational databases: {LGPS} and {LEOPARD},'' \emph{PDS}, vol. 20(3), pp.
  292--299, 2011.

\bibitem{Zorych2011}
I.~Zorych, D.~Madigan, P.~Ryan, and A.~Bate, ``Disproportionality methods for
  pharmacovigilance in longitudinal observational databases.'' \emph{Stat
  Methods Med Res}, vol. 0(0), pp. 1--18, 2011.

\bibitem{Jin2006}
H.~Jin, J.~Chen, C.~Kelman, H.~He, D.~McAullay, and C.~M. O'Keefe, ``Mining
  unexpected associations for signalling potential adverse drug reactions from
  administrative health databases,'' \emph{{PAKDD}}, pp. 867--876, 2006.

\bibitem{Jin2010}
H.~W. Jin, J.~Chen, H.~He, C.~Kelman, D.~McAullay, and C.~M. Okeefe,
  ``Signaling potential adverse drug reactions from administrative health
  databases,'' \emph{{IEEE} Transactions on knowledge and data engineering},
  vol. 22(6), pp. 839--853, 2010.

\bibitem{Ji2012}
Y.~Ji, H.~Y. anf John~Tran, P.~Dews, A.~Mansour, and R.~M. Massanari, ``A
  method for mining infrequent causal associations and its application in
  finding adverse drug reaction signal pairs,'' \emph{{IEEE} Transactions on
  knowledge and data engineering}, 2012.

\bibitem{Reps2013}
J.~Reps, J.~M. Garibaldi, U.~Aickelin, D.~Soria, J.~E. Gibson, and R.~B.
  Hubbard, ``Comparing data-mining algorithms developed for longitudinal
  observational databases.'' in \emph{{Computational Intelligence (UKCI), 2012
  12th UK Workshop on}}.\hskip 1em plus 0.5em minus 0.4em\relax Edinburgh:
  Heriot-Watt University, Sept. 2012, pp. 1--8.

\bibitem{Lewis2005}
J.~D. Lewis, W.~B. Bilker, R.~B. Weinstein, and B.~L. Strom, ``The relationship
  between time since registration and measured incidence rates in the general
  practice research database,'' \emph{Pharmacoepidemiol Drug Saf}, vol. 14(7),
  pp. 443--451, 2005.

\bibitem{netdr2012}
\BIBentryALTinterwordspacing
netdoctor. (2012) Netdoctor.co.uk - the uk's leading independent health
  website. online (accessed 09/08/2012). NetDoctor. [Online]. Available:
  \url{http://www.netdoctor.co.uk}
\BIBentrySTDinterwordspacing

\bibitem{Ying2012}
Y.~Ying and P.~Li, ``Distance metric learning with eigenvalue optimization,''
  \emph{JMLR}, vol.~13, pp. 1--26, 2012.

\bibitem{Basu2002}
S.~Basu, A.~Banerjee, and R.~Mooney, ``Semi-supervised clustering by seeding,''
  \emph{Proceedings of the 19th International Conference on Machine Learning
  ({ICML}-2002), Sydney, Australia}, pp. 19--26, 2002.

\bibitem{Graham2005}
D.~J. Graham, D.~Campen, R.~Hui, M.~Spence, C.~Cheetham, G.~Levy, S.~S, and
  W.~A. Ray, ``Risk of acute myocardial infarction and sudden cardiac death in
  patients treated with cyclo-oxygenase 2 selective and nonselective
  nonsteroidal anti-inflammatory drugs: nested case control study,''
  \emph{Lancet}, vol. 365(9458), pp. 475--81, 2005.

\bibitem{Despres2005}
J.~P. Despr�s, A.~Golay, and L.~S. L, ``Effects of rimonabant on metabolic
  risk factors in overweight patients with dyslipidemia.'' \emph{N Engl J Med},
  vol. 353(20), pp. 2121--2134, 2005.

\bibitem{Christensen2007}
R.~Christensen, P.~K. Kristensen, E.~M. Bartels, H.~Bliddal, and A.~Astrup,
  ``Efficacy and safety of the weight-loss drug rimonabant: a meta-analysis of
  randomised trials,'' \emph{Lancet}, vol. 370(9600), pp. 1706--1713, 2007.

\bibitem{Carnovale2003}
A.~Carnovale, P.~Esposito, P.~Bassano, L.~Russo, and G.~Uomo,
  ``Enalapril-induced acute recurrent pancreatitis,'' \emph{Digestive and Liver
  Disease}, vol. 35(1), pp. 55--57, 2003.

\bibitem{Balani2008}
A.~R. Balani and J.~H. Grendell, ``Drug-induced pancreatitis: Incidence,
  management and prevention,'' \emph{Drug Safety}, vol. 31(10), pp. 823--837,
  2008.

\bibitem{Harirforoosh2009}
S.~Harirforoosh and F.~Jamali, ``Renal adverse effects of nonsteroidal
  anti-inflammatory drugs.'' \emph{Expert Opin Drug Saf}, vol. 8(6), pp.
  669--681, 2009.

\bibitem{Ali2011}
S.~Ali, J.~D. Pimentel, and C.~Ma, ``Naproxen-induced liver injury.''
  \emph{Hepatobiliary Pancreat Dis Int}, vol. 10(5), pp. 552--556, 2011.

\bibitem{Onder2004}
G.~Onder, F.~Pellicciotti, G.~Gambassi, and R.~Bernabei, ``Nsaid-related
  psychiatric adverse events: Who is at risk?'' \emph{Drugs}, vol. 64(23), pp.
  2619--2627, 2004.

\bibitem{drugs2012}
\BIBentryALTinterwordspacing
{Drugs.com}. (2012) Nifedipine side effects from drugs.com;
  [accessed:13/12/2012 ]. [Online]. Available:
  \url{http://www.drugs.com/sfx/nifedipine-side-effects.html}
\BIBentrySTDinterwordspacing

\bibitem{Linden2002}
P.~D. van~der Linden, M.~C. J.~M. Sturkenboom, R.~M.~C. Herings, H.~G.~M.
  Leufkens, and B.~H.~C. Stricker, ``Fluoroquinolones and risk of achilles
  tendon disorders: case-control study,'' \emph{{BMJ}}, vol. 324, p. 1306,
  2002.

\bibitem{Lee1989}
C.~H. Lee and C.~J. Lee, ``Thrombocytopenia -- a rare but potentially serious
  side effect of initial daily and interrupted use of rifampicin,''
  \emph{{CHEST}}, vol. 96(1), pp. 202--203, 1989.

\bibitem{Chan1999}
N.~N. Chan, H.~P.~S. Brian, and M.~D. Feher, ``Metformin-associated lactic
  acidosis: a rare or very rare clinical entity?'' \emph{Diabetic Medicine},
  vol. 16(4), pp. 273--281, 1999.

\end{thebibliography}

%








\end{document}